\documentclass{llncs} 
\usepackage{llncsdoc}
\usepackage{hyperref}
\usepackage{url}
\usepackage{amsfonts,amscd,amssymb,amsmath}
\usepackage{bm}
\usepackage{bbm} 
\usepackage{color}

\usepackage{amsthm}

\newtheorem{thm}{Theorem}
\newtheorem{prop}{Proposition}

\newtheorem*{assumption}{Assumption}

\newcommand{\mc}{\mathcal}
\newcommand{\mb}{\mathbf}
\newcommand{\bb}{\mathbb}
\newcommand{\wh}{\widehat}

\title{Learning From Non-iid Data: Fast Rates for the One-vs-All Multiclass Plug-in Classifiers}

\author{Vu Dinh\inst{1}$^,$\thanks{These authors contributed equally to this work.} \and
Lam Si Tung Ho\inst{2}$^{,\star}$ \and
Nguyen Viet Cuong\inst{3} \and \\
Duy Nguyen\inst{4} \and
Binh T. Nguyen\inst{5}}
\institute{Department of Mathematics, Purdue University \\ 
\email{vdinh@math.purdue.edu} 
\and
Department of Human Genetics, University of California, Los Angeles \\
\email{lamho@ucla.edu} 
\and
Department of Computer Science, National University of Singapore \\
\email{nvcuong@comp.nus.edu.sg}
\and
Department of Statistics, University of Wisconsin-Madison \\
\email{dnguyen@stat.wisc.edu}
\and
Department of Computer Science, University of Science, Vietnam \\
\email{ngtbinh@hcmus.edu.vn}
}

\newcommand{\LP}{\,\text{LP}}

\begin{document} 
 
\maketitle
\setcounter{footnote}{0}

\vspace{-6mm}
\begin{abstract}
We prove new fast learning rates for the one-vs-all multiclass plug-in classifiers trained either from exponentially strongly mixing data or from data generated by a converging drifting distribution. These are two typical scenarios where training data are not iid.
The learning rates are obtained under a multiclass version of Tsybakov's margin assumption, 
a type of low-noise assumption, and do not depend on the number of classes. Our results are general and include a previous result for binary-class plug-in classifiers with iid data as a special case. In contrast to previous works for least squares SVMs under the binary-class setting, our results retain the optimal learning rate in the iid case.

\end{abstract}

\vspace{-6mm} 
\section{Introduction}
\vspace{-2mm}
Fast learning of plug-in classifiers from low-noise data has recently gained much attention \cite{audibert2007fast,kohler2007ontherate,monnier2012classification,minsker2012plug}. The first fast/super-fast learning rates\footnote{Fast learning rate means the trained classifier converges with rate faster than $n^{-1/2}$,
while super-fast learning rate means the trained classifier converges with rate faster than $n^{-1}$.}
for the plug-in classifiers were proven by Audibert and Tsybakov \cite{audibert2007fast} under the Tsybakov's margin assumption \cite{tsybakov2004optimal}, which is a type of low-noise condition.
Their plug-in classifiers employ the local polynomial estimator to estimate the conditional probability of a label $Y$ given an observation $X$ and use it in the plug-in rule.
Subsequently, Kohler and Krzyzak \cite{kohler2007ontherate} proved the fast learning rate for plug-in classifiers with a relaxed condition on the density of $X$
and investigated the use of kernel, partitioning, and nearest neighbor estimators instead of the local polynomial estimator.
Monnier \cite{monnier2012classification} suggested to use local multi-resolution projections to estimate the conditional probability of $Y$
and proved the super-fast rates of the corresponding plug-in classifier under the same margin assumption.
Fast rates for plug-in classifiers were also achieved in the active learning setting \cite{minsker2012plug}.

Nevertheless, these previous analyses of plug-in classifiers typically focus on the binary-class setting with iid (independent and identically distributed) data assumption.
This is a limitation of the current theory for plug-in classifiers since (1) many classification problems are multiclass in nature and (2) data may also violate the iid data assumption in practice.
In this paper, we contribute to the theoretical understandings of plug-in classifiers by proving novel fast learning rates of a multiclass plug-in classifier trained from non-iid data.
In particular, we prove that the multiclass plug-in classifier constructed using the \emph{one-vs-all method} can achieve fast learning rates, or even super-fast rates,
with the following two types of non-iid training data: data generated from an \emph{exponentially strongly mixing sequence} and data generated from a \emph{converging drifting distribution}.
To the best of our knowledge, this is the first result that proves fast learning rates for multiclass classifiers with non-iid data. Moreover, these learning rates do not depend on the number of classes.

Our results assume a multiclass version of Tsybakov's margin assumption.
In the multiclass setting, this assumption states that the events in which the most probable label of an example is ambiguous with the second most probable label have small probabilities.
This margin assumption was previously considered in the analyses of multiclass empirical risk minimization (ERM) classifiers with iid data \cite{Zhang04statistical}
and in the context of active learning with cost-sensitive multiclass classifiers \cite{agarwal2013selective}.
%
Our results are natural generalizations for both the binary-class and the iid data settings.
As special cases of our results, we can obtain fast learning rates for the one-vs-all multiclass plug-in classifiers in the iid data setting 
and the fast learning rates for the binary-class plug-in classifiers in the non-iid data setting.
Our results can also be used to obtain the previous fast learning rates \cite{audibert2007fast} for the binary-class plug-in classifiers in the iid data setting. 

In terms of theory, the extension from binary class to multiclass problem is usually not trivial
and depends greatly on the choice of the multiclass classifiers.
In this paper, our results show that this extension can be achieved with plug-in classifiers and the one-vs-all method. 
The one-vs-all method is a practical way to construct a multiclass classifier using binary-class classification \cite{rifkin2004defense}.
This method trains a model for each class by converting multiclass data into binary-class data and then combines them into a multiclass classifier.

Our paper considers two types of non-iid data.
Exponentially strongly mixing data is a typical case of identically but not independently distributed data.
Fast learning from exponentially strongly mixing data has been previously analyzed for least squares support vector machines (LS-SVMs) \cite{steinwart2009fast,Hang2014184} and ERM classifiers \cite{Hang2014184}.
On the other hand, data generated from a drifting distribution (or drifting concept) is an example of independently but not identically distributed data.
Some concept drifting scenarios and learning bounds were previously investigated in \cite{bartlett1992learning,long1999complexity,barve1996complexity,mohri2012new}.
In this paper, we consider the scenario where the parameters of the distributions generating the training data converge uniformly to those of the test distribution with some polynomial rate.

We note that even though LS-SVMs can be applied to solve a classification problem with binary data, 
the previous results for LS-SVMs cannot retain the optimal rate in the iid case \cite{steinwart2009fast,Hang2014184}.
In contrast, our results in this paper still retain the optimal learning rate for the H\"older class in the iid case.
Besides, the results for drifting concepts can also achieve this optimal rate.
Other works that are also related to our paper include the analyses of fast learning rates for binary SVMs and multiclass SVMs with iid data \cite{steinwart2007fast,Shen07general}
and for the Gibbs estimator with $\phi$-mixing data \cite{pierre2014prediction}.

\vspace{-2.5mm}
\section{Preliminaries}

\vspace{-1.5mm}
\subsection{Settings}

Let $\{ (X_i,Y_i) \}_{i=1}^n$ be the labeled training data where ${ X_i \in \bb{R}^d }$ and $Y_i \in \{1, 2, \allowbreak \ldots, m \}$ for all $i$.
In the data, $X_i$ is an observation and $Y_i$ is the label of $X_i$.
The binary-class case corresponds to $m = 2$, while the multiclass case corresponds to $m > 2$.
For now we do not specify how $\{ (X_i,Y_i) \}_{i=1}^n$ are generated,
but we assume that test data are drawn iid from an unknown distribution $\mb{P}$ on $\bb{R}^d \times \{1,2,\ldots, m\}$.
In Section \ref{sec:strongly-mixing} and \ref{sec:concept-drift}, we will respectively consider two cases where 
the training data $\{ (X_i,Y_i) \}_{i=1}^n$ are generated from an exponentially strongly mixing sequence with stationary distribution $\mb{P}$
and where $\{ (X_i,Y_i) \}_{i=1}^n$ are generated from a drifting distribution with the limit distribution $\mb{P}$.
The case where training data are generated iid from $\mb{P}$ is a special case of these settings.

Given the training data, our aim is to find a classification rule $f : \bb{R}^d \rightarrow \{1, 2, \ldots, m \}$ whose risk is as small as possible.
The risk of a classifier $f$ is defined as $R(f) \triangleq \mb{P}(Y \ne f(X))$.
One minimizer of the above risk is the Bayes classifier $f^*(X) \triangleq \arg\max_j \eta_j (X)$,
where ${ \eta_j (X) \triangleq \mb{P}(Y=j|X) }$ for all $j \in \{ 1, 2, \ldots, m \}$.
For any classifier $\wh{f}_n$ trained from the training data, it is common to characterize its accuracy via the excess risk $\mc{E}(\wh{f}_n) \triangleq \mb{E}R(\wh{f}_n) - R(f^*)$,
where the expectation is with respect to the randomness of the training data.
A small excess risk for $\wh{f}_n$ is thus desirable as the classifier will perform close to the optimal classifier $f^*$ on average.

For any classifier $f$, we write $\eta_f (X)$ as an abbreviation for $\eta_{f(X)}(X)$, which is the value of the function $\eta_{f(X)}$ at $X$.
Let $\mb{1}_{\{ \cdot \}}$ be the indicator function.
The following proposition gives a property of the excess risk in the multiclass setting.
This proposition will be used to prove the theorems in the subsequent sections.

\vspace{-1.5mm}
\begin{prop}
\label{prop:excess-risk}
For any classifier $\wh f_n$, we have $\mc{E}(\wh{f}_n) = \mb{E}\left[\eta_{f^*}(X)-\eta_{\wh f_n}(X) \right]$,
where the expectation is with respect to the randomness of both the training data and the testing example $X$.
\end{prop}

\vspace{-2.5mm}
\begin{proof}
$R(\wh f_n)-R(f^*)$
\begin{eqnarray*}
&=& \mb{P}( Y \ne \wh f_n (X) ) - \mb{P}(Y \ne f^*(X)) \,\,\,=\,\,\, \mb{P}(Y = f^*(X)) - \mb{P}(Y = \wh f_n(X)) \\
&=& \mb{E}_{X,Y} \left [ \mb{1}_{\{Y = f^*(X)\}} - \mb{1}_{\{Y = \wh f_n(X)\}} \right] = \mb{E}_X \left [ \mb{E}_Y \left [ \mb{1}_{\{Y = f^*(X)\}} - \mb{1}_{\{Y = \wh f_n(X)\}} \big | X \right ]\right ] \\
&=& \mb{E}_X \left[ \sum_{j=1}^m {\eta_j (X) \left( \mb{1}_{\{ f^*(X) = j \}} - \mb{1}_{\{ \wh f_n(X) = j \}} \right) }\right]
\,\,=\,\, \mb{E}_X \left[ \eta_{f^*}(X) - \eta_{\wh f_n}(X) \right].
\end{eqnarray*}
Thus, $\mc{E}(\wh{f}_n) = \mb{E}\left[\eta_{f^*}(X)-\eta_{\wh f_n}(X) \right]$.
\end{proof}

Following the settings for the binary-class case \cite{audibert2007fast}, we assume the following H\"{o}lder assumption: 
all the functions $\eta_j$'s are in the H\"{o}lder class $\Sigma(\beta, L, \bb{R}^d)$.
We also assume that the marginal distribution $\mb{P}_X$ of $X$ satisfies the strong density assumption.
The definition of H\"{o}lder classes and the strong density assumption are briefly introduced below by using the notations in \cite{audibert2007fast}.

For $\beta > 0$ and $L > 0$, the H\"{o}lder class $\Sigma(\beta, L, \bb{R}^d)$ is the set of all functions $g : \bb{R}^d \rightarrow \bb{R}$ that are $\lfloor \beta \rfloor$ times continuously differentiable,
and for any $x, x' \in \bb{R}^d$, we have $|g(x') - g_x(x')| \leq L ||x-x'||^\beta$,
where $||\cdot||$ is the Euclidean norm and $g_x$ is the $\lfloor \beta \rfloor^{th}$-degree Taylor polynomial of $g$ at $x$.
The definition of $g_x$ can be found in Section 2 of \cite{audibert2007fast}.

Fix $c_0, r_0 > 0$ and $0 < \mu_{\text{min}} < \mu_{\text{max}} < \infty$, and fix a compact set $\mc{C} \subset \bb{R}^d$. 
The marginal $\mb{P}_X$ satisfies the strong density assumption if it is supported on a compact $(c_0,r_0)$-regular set $A \subseteq \mc{C}$
and its density $\mu$ (w.r.t. the Lebesgue measure) satisfies:
$\mu_{\text{min}} \leq \mu(x) \leq \mu_{\text{max}}$ for $x \in A$ and $\mu(x) = 0$ otherwise.
In this definition, a set $A$ is $(c_0,r_0)$-regular if $\bm{\lambda} [A \cap B(x,r)] \geq c_0 \bm{\lambda}[B(x,r)]$ for all $0 < r \leq r_0$ and $x \in A$,
where $\bm{\lambda}$ is the Lebesgue measure and $B(x,r)$ is the Euclidean ball in $\bb{R}^d$ with center $x$ and radius $r$.

\subsection{Margin Assumption for Multiclass Setting}
\label{sec:margin-assumption}
As in the binary-class case, fast learning rates for the multiclass plug-in classifier can be obtained under an assumption similar to Tsybakov's margin assumption \cite{tsybakov2004optimal}.
In particular, we assume that the conditional probabilities $\eta_j$'s satisfy the following margin assumption, which is an extension of Tsybakov's margin assumption to the multiclass setting.
This is a form of low noise assumption and was also considered in the context of active learning to analyze the learning rate of cost-sensitive multiclass classifiers \cite{agarwal2013selective}.

\begin{assumption}[Margin Assumption]
There exist constants $C_0 > 0$ and $\alpha \geq 0$ such that for all $t > 0$,
\[ \mb{P}_X( \eta_{(1)}(X) - \eta_{(2)}(X) \leq t) \leq C_0 t^\alpha \]
where $\eta_{(1)}(X)$ and $\eta_{(2)}(X)$ are the largest and second largest conditional probabilities among all the $\eta_j(X)$'s.
\end{assumption}

\section{The One-vs-All Multiclass Plug-in Classifier}
\label{sec:1-vs-all}
We now introduce the one-vs-all multiclass plug-in classifier which we will analyze in this paper.
Let $\wh \eta_n (X) = (\wh \eta_{n,1} (X), \wh \eta_{n,2} (X), \ldots, \wh \eta_{n,m} (X))$ be an $m$-dimensional function where $\wh \eta_{n,j}$ is a nonparametric estimator of $\eta_j$ from the training data.
The corresponding multiclass plug-in classifier $\wh f_n$ predicts the label of an observation $X$ by
\[
\wh f_n (X) = \arg\max_j {\wh \eta_{n,j} (X)}.
\]
In this paper, we consider plug-in classifiers where $\wh \eta_{n,j}$'s are estimated using the one-vs-all method and the local polynomial regression function as follows.
For each class $j \in \{ 1, 2, \ldots, m \}$, we first convert the training data $\{ (X_i,Y_i) \}_{i=1}^n$ to binary class by considering all $(X_i,Y_i)$'s such that $Y_i \ne j$ as negative (label 0)
and those such that $Y_i = j$ as positive (label 1).
Then we construct a local polynomial regression function $\wh \eta^{\LP}_{n,j} (x)$ of order $\lfloor \beta \rfloor$ 
with some appropriate bandwidth $h > 0$ and kernel $K$ from the new binary-class training data (see Section 2 of \cite{audibert2007fast} for the definition of local polynomial regression functions).
The estimator $\wh \eta_{n,j}$ can now be defined as
\[
\wh \eta_{n,j}(x) \triangleq
\begin{cases} 
0 & \mbox{if } \wh \eta^{\LP}_{n,j}(x) \leq 0 \\ 
\wh \eta^{\LP}_{n,j}(x) & \mbox{if } 0< \wh \eta^{\LP}_{n,j}(x) < 1 \\
1 & \mbox{if } \wh \eta^{\LP}_{n,j}(x) \geq 1
\end{cases}.
\]
In order to prove the fast rates for the multiclass plug-in classifier, the bandwidth $h$ and the kernel $K$ of the local polynomial regression function have to be chosen carefully.
Specifically, $K$ has to satisfy the following assumptions, which are similar to those in \cite{audibert2007fast}:
\[
\exists c > 0 \text{ such that for all } x \in \bb{R}^d, \text{ we have } K(x) \geq c \mb{1}_{ \{ ||x|| \leq c \} },
\]
\[
\int_{\bb{R}^d} K(u) du = 1, \,\,\, \sup_{u \in \bb{R}^d}(1 + ||u||^{2 \beta}) K(u) < \infty, \,\,\, \text{and } \int_{\bb{R}^d}(1 + ||u||^{4 \beta}) K^2(u) du < \infty.
\]
Note that Gaussian kernels satisfy these conditions.
The conditions for the bandwidth $h$ will be given in Section \ref{sec:strongly-mixing} and \ref{sec:concept-drift}.

\section{Fast Learning For Exponentially Strongly Mixing Data}
\label{sec:strongly-mixing}
In this section, we consider the case where training data are generated from an exponentially strongly mixing sequence \cite{steinwart2009fast,modha1996minimum}.
Let $Z_i = (X_i, Y_i)$ for all $i$.
Assume that $\{ Z_i \}_{i=1}^\infty$ is a stationary sequence of random variables on $\bb{R}^d \times \{ 1, 2, \ldots, m \}$ with stationary distribution $\mb{P}$. 
That is, $\mb{P}$ is the marginal distribution of any random variable in the sequence.
For all $k \geq 1$, we define the $\boldsymbol \alpha$-mixing coefficients \cite{steinwart2009fast}:
\[
\boldsymbol \alpha(k) \triangleq \sup_{A_1 \in \sigma_1^t, A_2 \in \sigma_{t+k}^\infty, t \geq 1}{|\mb{P}(A_1 \cap A_2) - \mb{P}(A_1)\mb{P}(A_2)|}
\]
where $\sigma_a^b$ is the $\sigma$-algebra generated by $\{ Z_i \}_{i=a}^b$.
The sequence $\{ Z_i \}_{i=1}^\infty$ is exponentially strongly mixing if there exist positive constants $C_1$, $C_2$ and $C_3$ such that for every $k \geq 1$, we have
\begin{equation}
\label{eqn:mixing}
\boldsymbol \alpha(k) \leq C_1 \exp(-C_2 k^{C_3}).
\end{equation}
We now state some key lemmas for proving the convergence rate of the multiclass plug-in classifier in this setting.
Let $n_e \triangleq \left \lfloor \frac{n}{\lceil \{8n/C_2\}^{1/(C_3 + 1)} \rceil} \right \rfloor$ be the effective sample size.
The following lemma is a direct consequence of Bernstein inequality for an exponentially strongly mixing sequence \cite{modha1996minimum}.

\begin{lemma}
\label{lem_Bernstein}
Let $\{ Z_i \}_{i=1}^\infty$ be an exponentially strongly mixing sequence and $\phi$ be a real-valued Borel measurable function.
Denote $W_i = \phi(Z_i)$ for all $i \geq 1$.
Assume that $|W_1| \leq C$ almost surely and $\mb E[W_1] = 0$.
Then for all $n \geq 1$ and $\epsilon >0$, we have
\[
\mb P^{\otimes n} \left( \left |\frac{1}{n} \sum_{i=1}^n W_i \right | \geq \epsilon \right) \leq 2(1+4e^{-2} C_1) \exp \left( - \frac{\epsilon^2 n_e}{2\mb E|W_1|^2 + 2 \epsilon C / 3} \right),
\]
where $\mb P^{\otimes n}$ is the joint distribution of $\{Z_i\}_{i=1}^n$ and $C_1$ is the constant in Eq. \eqref{eqn:mixing}.
\end{lemma}

The next lemma is about the convergence rate of the local polynomial regression functions using the one-vs-all method.
The proof for this lemma is given in Section \ref{sec:lem_mixing_proof}.

\begin{lemma}
\label{lem_mixing}
Let $\beta$, $r_0$, and $c$ be the constants in the H\"{o}lder assumption, the strong density assumption, and the assumption for the kernel K respectively.
Then there exist constants $C_4, C_5, C_6 > 0$ such that for all $\delta > 0$, all bandwidth $h$ satisfying $C_6 h^\beta < \delta$  and $0<h\leq r_0/c$, 
all $j \in \{1,2, \ldots, m\}$ and $n \geq 1$, we have
\[
\mb P^{\otimes n}(|\wh \eta_{n,j}(x) - \eta_j(x)| \geq \delta) \leq C_4 \, \exp(- C_5 n_e h^d\delta^2)
\]
for almost surely $x$ with respect to $\mb{P}_X$, where $d$ is the dimension of the observations (inputs).
\end{lemma}

Given the above convergence rate of the local polynomial regression functions, 
Lemma \ref{lem:rate} below gives the convergence rate of the excess risk of the one-vs-all multiclass plug-in classifier.
The proof for this lemma is given in Section \ref{lem:rate_proof}.

\begin{lemma}
\label{lem:rate}
Let $\alpha$ be the constant in the margin assumption.
Assume that there exist $C_4, C_5 > 0$ such that $\mb P^{\otimes n}(|\wh \eta_{n,j}(x) - \eta_j(x)| \geq \delta) \leq C_4 \, \exp(-C_5 a_n \delta^2)$
for almost surely $x$ with respect to $\mb P_X$, and for all $j \in \{1,2, \ldots, m\}$, $\delta > 0$.
Then there exists $C_7 > 0$ such that for all $n \geq 1$,
\[
\mc{E}(\wh{f}_n) = \mb{E}R(\wh f_n) - R(f^*) \leq C_7 a_n^{-(1+\alpha)/2}.
\]
\end{lemma}

Using Lemma \ref{lem_mixing} and \ref{lem:rate}, we can obtain the following theorem about the convergence rate of the one-vs-all multiclass plug-in classifier
when training data are exponentially strongly mixing.
This theorem is a direct consequence of Lemma \ref{lem_mixing} and \ref{lem:rate} with $h=n_e^{-1/(2 \beta +d)}$ and $a_n=n_e^{2\beta/(2 \beta +d)}$.

\begin{thm}
\label{thm_fastrate}
Let $\alpha$ and $\beta$ be the constants in the margin assumption and the H\"{o}lder assumption respectively,
and let $d$ be the dimension of the observations.
Let $\wh f_n$ be the one-vs-all multiclass plug-in classifier with bandwidth $h=n_e^{-1/(2 \beta +d)}$ that is trained from an exponentially strongly mixing sequence.
Then there exists some constant $C_8 >0$ such that for all $n$ large enough that satisfies $0< n_e^{-1/(2 \beta +d)}\leq r_0/c$, we have
\[
\mc{E}(\wh{f}_n) = \mb{E} R(\wh f_n) - R(f^*) \le C_8 n_e^{-\beta(1+\alpha)/(2 \beta+d)}.
\]
\end{thm}

The convergence rate in Theorem \ref{thm_fastrate} is expressed in terms of the effective sample size $n_e$ rather than the sample size $n$
since learning with dependent data typically requires more data to achieve the same level of accuracy as learning with independent data (see e.g., \cite{steinwart2009fast,cuong2013generalization,ane2008analysis}).
However, Theorem \ref{thm_fastrate} still implies the fast rate for the one-vs-all multiclass plug-in classifier in terms of the sample size $n$.
Indeed, the rate in the theorem can be rewritten as $O(n^{-\frac{\beta(1+\alpha)}{2\beta + d} \cdot \frac{C_3}{C_3+1}})$, so the fast learning rate is achieved when $2(\alpha-1/C_3)\beta > (1+1/C_3)d$
and the super-fast learning rate is achieved when $(\alpha - 1 - 2/C_3)\beta>d(1+1/C_3)$.

\vspace{-2mm}
\section{Fast Learning From a Drifting Concept}
\label{sec:concept-drift}
\vspace{-1mm}

In this section, we consider the case where training data are generated from a drifting concept that converges to the test distribution $\mb{P}$.
Unlike the setting in Section \ref{sec:strongly-mixing} where the training data form a stationary sequence of random variables, 
the setting in this section may include training data that are not stationary.
Formally, we assume the training data $\{ Z_i \}_{i=1}^n = \{ (X_i,Y_i) \}_{i=1}^n$ are generated as follows.
The observations $X_i$ are generated iid from the marginal distribution $\mb P_X$ satisfying the strong density assumption.
For each $i \ge 1$, the label $Y_i$ of $X_i$ is generated from a categorical distribution on $\{1,2,\ldots,m\}$ with parameters $\eta^i (X_i) \triangleq (\eta^i_1 (X_i), \eta^i_2 (X_i), \ldots, \eta^i_m (X_i))$.
That is, the probability of $Y_i = j$ conditioned on $X_i$ is $\eta^i_j (X_i)$, for all $j \in \{1,2,\ldots,m\}$.

Note that from our setting, the training data are independent but not identically distributed.
To prove the convergence rate of the multiclass plug-in classifier, we assume that 
$\|\eta^n_j - \eta_j \|_\infty \triangleq \sup_{x \in \bb{R}^d} |\eta^n_j(x) - \eta_j(x)| = O(n^{-(\beta+d)/(2\beta + d)})$ for all $j$,
i.e., $\eta^n_j$ converges uniformly to the label distribution $\eta_j$ of test data with rate $O(n^{-(\beta+d)/(2\beta + d)})$.
We now state some useful lemmas for proving our result.
The following lemma is a Bernstein inequality for the type of data considered in this section \cite{yurinskiui1976exponential}.

\vspace{-1mm}
\begin{lemma}
\label{lem_Bernstein_ind}
Let $\{ W_i \}_{i=1}^n$ be an independent sequence of random variables.
For all $i \ge 1$ and $l > 2$, assume $\mb E W_i = 0$, $\mb E |W_i|^2 = b_i$, and $\mb E|W_i|^l \leq b_i H^{l-2} l! /2$ for some constant $H > 0$.
Let $B_n \triangleq \sum_{i=1}^n{b_i}$.
Then for all $n \ge 1$ and $\epsilon > 0$, we have
\[
\mb P^{\otimes n} \left( \left| \sum_{i=1}^n{W_i} \right| \ge \epsilon \right) \leq 2 \exp \left( - \frac{\epsilon^2}{2(B_n + H \epsilon)} \right),
\]
where $\mb P^{\otimes n}$ is the joint distribution of $\{ W_i \}_{i=1}^n$.
\end{lemma}

The next lemma states the convergence rate of the local polynomial regression functions in this setting.
The proof for this lemma is given in Section \ref{sec:lem:approX_jnd_proof}.
Note that the constants in this section may be different from those in Section \ref{sec:strongly-mixing}.

\vspace{-1mm}
\begin{lemma}
\label{lem:approX_jnd}
Let $\beta$, $r_0$, and $c$ be the constants in the H\"{o}lder assumption, the strong density assumption, and the assumption for the kernel K respectively.
Let $\wh \eta_{n,j}$ be the estimator of $\eta_j$ estimated using the local polynomial regression function with ${ h = n^{- 1/(2\beta+d)} }$.
If ${ \|\eta^n_j - \eta_j\|_\infty = O(n^{-(\beta+d)/(2\beta + d)}) }$ for all $j$,
then there exist constants $C_4, C_5, C_6 > 0$ such that for all $\delta > 0$, all $n$ satisfying $C_6 n^{- \beta/(2\beta+d)} < \delta < 1$  and $0< n^{- 1/(2\beta+d)} \leq r_0/c$, 
and all $j \in \{1,2, \ldots, m\}$, we have
\[
\mb P^{\otimes n}(|\wh \eta_{n,j}(x) - \eta_j(x)| \geq \delta) \leq C_4 \, \exp(- C_5 n^{2\beta/(2\beta+d)} \delta^2)
\]
for almost surely $x$ with respect to $\mb{P}_X$, where $d$ is the dimension of the observations.
\end{lemma}

Note that Lemma \ref{lem:rate} still holds in this setting.
Thus, we can obtain Theorem \ref{thm_fastrate_drift} below about the convergence rate of the one-vs-all multiclass plug-in classifier 
when training data are generated from a drifting concept converging uniformly to the test distribution.
This theorem is a direct consequence of Lemma \ref{lem:rate} and \ref{lem:approX_jnd} with $a_n=n^{2\beta/(2 \beta +d)}$.
We note that the convergence rate in Theorem \ref{thm_fastrate_drift} is fast when $\alpha \beta > d/2$ and is super-fast when $(\alpha-1)\beta>d$.

\vspace{-1mm}
\begin{thm}
\label{thm_fastrate_drift}
Let $\alpha$ and $\beta$ be the constants in the margin assumption and the H\"{o}lder assumption respectively,
and let $d$ be the dimension of the observations.
Let $\wh f_n$ be the one-vs-all multiclass plug-in classifier with bandwidth $h=n^{-1/(2 \beta +d)}$ 
that is trained from data generated from a drifting concept converging uniformly to the test distribution.
Then there exists some constant $C_8 >0$ such that for all $n$ large enough that satisfies $0< n^{-1/(2 \beta +d)}\leq r_0/c$, we have
\[
\mc{E}(\wh{f}_n) = \mb{E} R(\wh f_n) - R(f^*) \le C_8 n^{-\beta(1+\alpha)/(2 \beta+d)}.
\]
\end{thm}

\vspace{-5mm}
\section{Remarks}
\vspace{-2mm}

The rates in Theorem \ref{thm_fastrate} and \ref{thm_fastrate_drift} do not depend on the number of classes $m$.
They are both generalizations of the previous result for binary-class plug-in classifiers with iid data \cite{audibert2007fast}.
More specifically, $C_3 = +\infty$ in the case of iid data, thus we have $n_e = n$ and the data distribution also satisfies the condition in Theorem \ref{thm_fastrate_drift}.
Hence, we can obtain the same result as in \cite{audibert2007fast}.

Another important remark is that our results for the one-vs-all multiclass plug-in classifiers retain the optimal rate $O(n^{-\beta(1+\alpha)/(2\beta + d)})$ for the H\"older class in the iid case \cite{audibert2007fast}
while the previous results in \cite{steinwart2009fast,Hang2014184} for LS-SVMs with smooth kernels do not (see Example 4.3 in \cite{Hang2014184}).
Besides, from Theorem \ref{thm_fastrate_drift}, the one-vs-all multiclass plug-in classifiers trained from a drifting concept can also achieve this optimal rate.
We note that for LS-SVMs with Gaussian kernels, Hang and Steinwart \cite{Hang2014184} proved that they can achieve the essentially optimal rate in the iid scenario (see Example 4.4 in \cite{Hang2014184}).
That is, their learning rate is $n^{\zeta}$ times of the optimal rate for any $\zeta > 0$. 
Although this rate is very close to the optimal rate, it is still slower than $\log n$ times of the optimal rate.\footnote{The 
optimal rates in Example 4.3 and 4.4 of \cite{Hang2014184} may not necessarily be the same as our optimal rate since Hang and Steinwart considered Sobolev space and Besov space instead of H\"older space.}

\vspace{-2mm}
\section{Technical Proofs}

\subsection{Proof of Lemma \ref{lem_mixing}}
\label{sec:lem_mixing_proof}

Fix $j \in \{ 1, \ldots, m \}$.
Let $Y'_i \triangleq \mb{1}_{\{ Y_i = j \}}$ be the binary class of $X_i$ constructed from the class $Y_i$ using the one-vs-all method in Section \ref{sec:1-vs-all}.
By definition of $\eta_j$, note that $\mathbf{P}[Y'_i = 1 | X_i] = \eta_j(X_i)$.
Let $\mu$ be the density of $\mb P_X$.
We consider the matrix $\mb B \triangleq (B_{s_1,s_2})_{|s_1|,|s_2| \leq \lfloor \beta \rfloor}$ with the elements
${ B_{s_1,s_2} \triangleq \int_{\bb R^d}{u^{s_1+s_2}K(u) \mu(x + h u)du} }$,
and the matrix $\wh {\mb B} \triangleq (\wh B_{s_1,s_2})_{|s_1|,|s_2| \leq \lfloor \beta \rfloor}$ with the elements
$\wh B_{s_1,s_2} \triangleq \linebreak \frac{1}{n h^d} \sum_{i=1}^n (\frac{X_i - x}{h})^{s_1+s_2} K(\frac{X_i - x}{h})$,
where $s_1, s_2$ are multi-indices in $\bb N^d$ (see Section 2 of \cite{audibert2007fast} for details on multi-index).
Let $\lambda_{\mb B}$ be the smallest eigenvalue of $\mb B$. 
Then, there exists a constant $c_1$ such that ${ \lambda_{\mb B} \geq c_1 > 0 }$ (see Eq. (6.2) in \cite{audibert2007fast}).

Fix $s_1$ and $s_2$. For any $i = 1, 2, \ldots, n$, we define
\[
T_i \triangleq \frac{1}{h^d} \left( \frac{X_i - x}{h} \right)^{s_1+s_2} K \left( \frac{X_i - x}{h} \right) - \int_{\bb R^d}{u^{s_1+s_2}K(u) \mu(x + h u) du}.
\]
It is easy to see that $\mb E[T_1] = 0$, $|T_1| \leq c_2 h^{-d}$, and $\mb E|T_1|^2 \leq c_3 h ^{-d}$ for some $c_2, c_3 > 0$.
By applying Lemma \ref{lem_Bernstein}, for any $\epsilon > 0$, we have
\begin{align*}
\mb P^{\otimes n}(|\wh B_{s_1,s_2} - B_{s_1,s_2}| \geq \epsilon) &= \mb P^{\otimes n} \left( \left| \frac{1}{n} \sum_{i=1}^n T_i \right| \geq \epsilon \right) \\
&\leq 2(1+4e^{-2}C_1) \exp \left( - \frac{\epsilon^2 n_e h^d}{2 c_3 + 2 \epsilon c_2 / 3} \right).
\end{align*}
Let $\lambda_{\wh {\mb B}}$ be the smallest eigenvalue of $\wh {\mb B}$.
From Eq. (6.1) in \cite{audibert2007fast}, we have
\[
\lambda_{\wh {\mb B}} \geq \lambda_{\mb B} - \sum_{|s_1|,|s_2| \leq \lfloor \beta \rfloor} |\wh B_{s_1,s_2} - B_{s_1,s_2}|.
\]
Let $M$ be the number of columns of $\wh B$. Then, there exists $c_4 > 0$ such that
\begin{eqnarray}
\label{eqn:bound1}
\mb P^{\otimes n}(\lambda_{\wh {\mb B}} \leq c_1/2) \leq 2(1+4e^{-2} C_1) M^2 \exp(-c_4 n_e h^d).
\end{eqnarray}
Let $\eta^x_j$ be the $\lfloor \beta \rfloor^{th}$-degree Taylor polynomial of $\eta_j$ at $x$.
Consider the vector ${ \mb a \triangleq (a_s)_{|s|\leq \lfloor \beta \rfloor} \in \mathbb{R}^M }$ where
$a_s \triangleq \frac{1}{n h^d} \sum_{i=1}^n {[Y'_i - \eta^x_j(X_i)] ( \frac{X_i - x}{h} )^s K ( \frac{X_i - x}{h} )}$.
Applying Eq. (6.5) in \cite{audibert2007fast} for ${ \lambda_{\wh {\mb B}} \geq c_1/2 }$, we have
\vspace{-1mm}
\begin{equation}
\label{eqn:bound2}
|\wh \eta_{n,j}(x) - \eta_j(x)| \leq |\wh \eta^{\LP}_{n,j}(x) - \eta_j(x)| \leq \lambda_{\wh {\mb B}}^{-1} M \max_s|a_s| \leq (2 M / c_1) \max_s|a_s|.
\end{equation}
\vspace{-2mm}
We also define: \qquad $\displaystyle T^{(s,1)}_i \triangleq \frac{1}{h^d} [Y'_i - \eta_j(X_i)] ( \frac{X_i - x}{h} )^s K ( \frac{X_i - x}{h} )$, and
\[ 
{\hskip 2cm} T^{(s,2)}_i \triangleq \frac{1}{h^d} [\eta_j(X_i) - \eta^x_j(X_i)] ( \frac{X_i - x}{h} )^s K ( \frac{X_i - x}{h} ).
\]
Note that $\mb{E}[T_1^{(s,1)}]=0$, $|T_1^{(s,1)}| \le c_5 h^{-d}$ and $\mb{E}|T_1^{(s,1)}|^2 \le c_6 h^{-d}$ for some $c_5, c_6 > 0$.
Similarly, $|T_1^{(s,2)}-\mb{E}T_1^{(s,2)}| \le c_7 h^{\beta-d}+ c_8 h^{\beta} \le c_9 h^{\beta - d}$ and $\mb{E} |T_1^{(s,2)}-\mb{E}T_1^{(s,2)}|^2 \le c_{10}h^{2\beta-d}$, for some $c_7$, $c_8$, $c_9$, $c_{10} > 0$.
Thus, by applying Lemma \ref{lem_Bernstein} again, for any $\epsilon_1, \epsilon_2 > 0$, we have
\[
\mb P^{\otimes n} \left( \left|\frac{1}{n}\sum_{i=1}^n { T_i^{(s,1)}}\right| \geq \epsilon_1 \right) \le 2(1+4e^{-2}C_1) \exp{\left( - \frac{\epsilon_1^2 n_e h^d}{2 c_6+ 2 c_5 \epsilon_1/3} \right)}, \text{ and}
\]
\[ 
\displaystyle \mb P^{\otimes n} ( | \frac{1}{n}\sum_{i=1}^n{ (T_i^{(s,2)} {\hskip -1mm} - \mb{E} T_i^{(s,2)}) } | \geq \epsilon_2 ) \le 2(1+4e^{-2}C_1) \exp{\left(\frac{-\epsilon_2^2 n_e h^d}{2 c_{10}  h^{2\beta} + 2 c_{9} h^{\beta}\epsilon_2/3}\right)}. 
\]
Moreover, $|\mb{E}T_1^{(s,2)}| \le c_{8} h^{\beta}$.
By choosing $h^\beta \leq c_1 \delta/(6 M c_{8})$, there exists $c_{11} > 0$ such that $\displaystyle \mb P^{\otimes n} \left ( |a_s| \geq \frac{c_1 \delta}{2M} \right )$
\begin{eqnarray}
&\leq& \mb P^{\otimes n} \left ( \left | \frac{1}{n} \sum_{i=1}^n{T_i^{(s,1)}} \right | \geq \frac{c_1 \delta}{6M} \right ) 
+ \mb P^{\otimes n} \left ( \left | \frac{1}{n} \sum_{i=1}^n{(T_i^{(s,2)}-\mb{E}T_i^{(s,2)})} \right | \geq \frac{c_1 \delta}{6M} \right ) \nonumber \\
&\leq& 4 (1+4e^{-2}C_1) \exp(-c_{11}n_e h^d \delta^2).
\label{eqn:bound3}
\end{eqnarray}
Let $C_6 = 6 M c_8 / c_1$. By \eqref{eqn:bound1}, \eqref{eqn:bound2}, and \eqref{eqn:bound3}, there exist $C_4, C_5 > 0$ such that
\begin{eqnarray*}
& & \mb P^{\otimes n}(|\wh \eta_{n,j}(x) - \eta_j(x)| \geq \delta) \\
&\leq& \mb P^{\otimes n}(\lambda_{\wh {\mb B}} \leq c_1/2) + \mb P^{\otimes n}(|\wh \eta_{n,j}(x) - \eta_j(x)| \geq \delta, \lambda_{\wh {\mb B}} > c_1/2) \\
&\leq& C_4 \exp(- C_5 n_e h^d \delta^2).
\end{eqnarray*}
Note that the constants $C_4, C_5, C_6$ can be modified so that they are the same for all $\delta$, $h$, $j$, and $n$.
Thus, Lemma \ref{lem_mixing} holds.

\subsection{Proof of Lemma \ref{lem:rate}}
\label{lem:rate_proof}

Since $\eta_{f^*}(x) -\eta_{\wh f_n}(x) \ge 0$ for all $x \in \bb R^d$, we denote, for any $\delta>0$,
\[
A_0 \triangleq \{x \in \bb R^d : \eta_{f^*}(x) - \eta_{\wh f_n}(x) \le \delta\}, \text{ and }
\]
\[
A_i \triangleq \{x \in \bb R^d : 2^{i-1}\delta<\eta_{f^*}(x) - \eta_{\wh f_n}(x) \le 2^i \delta\},  \text{ for } i\geq 1.
\]
By Proposition \ref{prop:excess-risk}, $\mb{E}R(\wh f_n) - R(f^*) = \mb{E} [(\eta_{f^*}(X)-\eta_{\wh f_n}(X)) \, \mb{1}_{\{\wh f_n(X)\ne f^*(X)\}}]$
\begin{eqnarray*}
&=& \sum_{i=0}^{\infty}{\mb{E}\left[(\eta_{f^*}(X)-\eta_{\wh f_n}(X)) \, \mb{1}_{\{\wh f_n(X)\ne f^*(X)\}} \, \mb{1}_{ \{ X \in A_i \} }\right]} \\
&\le& \delta \mb{P} \left( 0<\eta_{f^*}(X)-\eta_{\wh f_n}(X) \le \delta \right) \\
& & + \sum_{i=1}^{\infty}{\mb{E}\left[(\eta_{f^*}(X)-\eta_{\wh f_n}(X)) \, \mb{1}_{\{\wh f_n(X)\ne f^*(X)\}} \, \mb{1}_{ \{ X \in A_i \} }\right]}.
\end{eqnarray*}
Let ${\wh \eta}_{n, \wh f_n}(x)$ denote ${\wh \eta}_{n, \wh f_n(x)}(x)$.
For any $x$, since ${\wh \eta}_{n, \wh f_n}(x)$ is the largest among ${\wh \eta}_{n, j}(x)$'s, we have
${ \eta_{f^*}(x)-\eta_{\wh f_n}(x) \le |\eta_{f^*}(x) - {\wh \eta}_{n,f^*}(x)| + |{\wh \eta}_{n, \wh f_n}(x) -\eta_{\wh f_n}(x)| }$.
For any $i \ge 1$, we have
\begin{eqnarray*}
&&\mb{E}\left[(\eta_{f^*}(X)-\eta_{\wh f_n}(X)) \, \mb{1}_{\{\wh f_n(X)\ne f^*(X)\}} \, \mb{1}_{ \{ X \in A_i \}}\right]\\
&\le& 2^i \delta \, \mb{E}\left[ \mb{1}_{\{ |\eta_{f^*}(X) - {\wh \eta}_{n,f^*}(X)| + |{\wh \eta}_{n, \wh f_n}(X) -\eta_{\wh f_n}(X)| \ge 2^{i-1}\delta\}}~\mb{1}_{\{0<\eta_{f^*}(X)-\eta_{\wh f_n}(X)<2^i \delta\}}\right]\\
&\le& 2^i \delta \, \mb{E}_X [\mb{P}^{\otimes n}( |\eta_{f^*}(X) - {\wh \eta}_{n,f^*}(X)| + |{\wh \eta}_{n, \wh f_n}(X) -\eta_{\wh f_n}(X)| \ge 2^{i-1}\delta) \cdot \\
& & {\hskip 1.2cm} \mb{1}_{\{0<\eta_{f^*}(X)-\eta_{\wh f_n}(X)<2^i \delta\}} ]\\
&\le& c_1 2^i \delta \exp \left(- c_2 a_n(2^{i-2}\delta)^2\right) ~ \mb{P}_X (0<\eta_{f^*}(X)-\eta_{\wh f_n}(X)<2^i \delta),
\end{eqnarray*}
for some $c_1, c_2 > 0$.
We have $\mb{P}_X (0<\eta_{f^*}(X)-\eta_{\wh f_n}(X)<\delta) \le \linebreak \mb{P}_X [\eta_{f^*}(X)-\eta_{(2)}(X)<\delta]$,
and by the margin assumption, for all $t>0$, we get
$\mb{P}_X [\eta_{f^*}(X)-\eta_{(2)}(X)<t] \le C_0 t^{\alpha}$.
Therefore,
\vspace{-2mm}
\begin{eqnarray*}
& & \mb{E}\left[(\eta_{f^*}(X)-\eta_{\wh f_n}(X)) \, \mb{1}_{\{\wh f_n(X)\ne f^*(X)\}} \, \mb{1}_{ \{ X \in A_i \}}\right] \\
&\le& c_1 C_0 2^{i(\alpha+1)} \delta^{\alpha+1} \exp \left(-c_2 a_n(2^{i-2}\delta)^2\right).
\end{eqnarray*}
By choosing $\delta=a_n^{-1/2}$, there exists $C_7 > 0$ that does not depend on $n$ and
\begin{eqnarray*}
\mb{E}R(\wh f_n) - R(f^*) &\le& C_0 a_n^{-(\alpha+1)/2} + ~2 c_1 C_0 a_n^{-(\alpha+1)/2} \sum_{i\ge 1} 2^{i(\alpha+1)/2}\exp(-c_2 2^{2i-4}) \\
&\le& C_7 a_n^{-(\alpha+1)/2}.
\end{eqnarray*}

\vspace{-5mm}
\subsection{Proof of Lemma \ref{lem:approX_jnd}}
\label{sec:lem:approX_jnd_proof}

The proof for this lemma is essentially similar to the proof for Lemma \ref{lem_mixing} in Section \ref{sec:lem_mixing_proof},
except that we use the Bernstein inequality for iid random variables to bound $\mb P^{\otimes n}(|\wh B_{s_1,s_2} - B_{s_1,s_2}| \geq \epsilon)$
and thus obtain $\mb P^{\otimes n}(\lambda_{\wh {\mb B}} \leq c_1/2) \leq 2 M^2 \exp(-c_4 n h^d)$ as an analogy of Eq. \eqref{eqn:bound1} in Section \ref{sec:lem_mixing_proof}.
Besides, Eq. \eqref{eqn:bound2} can be obtained in the same way as in Section \ref{sec:lem_mixing_proof}.
To obtain the bound similar to Eq. \eqref{eqn:bound3}, we define
\begin{eqnarray*}
T^{(s,1)}_i &\triangleq& \frac{1}{h^d} [Y'_i - \eta^i_j(X_i)] (\frac{X_i - x}{h})^s K(\frac{X_i - x}{h}) \\
T^{(s,2)}_i &\triangleq& \frac{1}{h^d} [\eta^i_j(X_i) - \eta_j(X_i)] (\frac{X_i - x}{h})^s K(\frac{X_i - x}{h}) \\
T^{(s,3)}_i &\triangleq& \frac{1}{h^d} [\eta_j(X_i) - \eta^x_j(X_i)] (\frac{X_i - x}{h})^s K(\frac{X_i - x}{h}).
\end{eqnarray*}
Note that $\mb{E}[T_i^{(s,1)}] = 0$, $|T_i^{(s,1)}| \le c_5 h^{-d}$, and $\mb{E}|T_i^{(s,1)}|^2 \le c_6 h^{-d}$ for some $c_5, c_6 > 0$.
Thus, $\mb{E}|T_i^{(s,1)}|^l \le (c_5 h^{-d})^{l-2} \mb{E}|T_i^{(s,1)}|^2 \le H_1^{l-2} \mb{E}|T_i^{(s,1)}|^2 l!/2$,
where $H_1 \triangleq c_5 h^{-d}$ and $l > 2$.
Similarly, $|T_i^{(s,2)}-\mb{E}T_i^{(s,2)}| \le c_7 h^{- d}$ and $\text{Var}[T_i^{(s,2)}] \le c_8 h^{2-d}$ for some $c_7, c_8 > 0$.
Thus, $\mb{E}|T_i^{(s,2)}-\mb{E}T_i^{(s,2)}|^l \le H_2^{l-2} \text{Var}[T_i^{(s,2)}] l!/2$, for $H_2 \triangleq c_7 h^{- d}$ and $l > 2$.
Furthermore, $|T_i^{(s,3)}-\mb{E}T_i^{(s,3)}| \le c_9 h^{\beta-d}$ and $\text{Var}[T_i^{(s,3)}] \le c_{10}h^{2\beta-d}$ for some $c_9, c_{10} > 0$.
Hence, $\mb{E}|T_i^{(s,3)}-\mb{E}T_i^{(s,3)}|^l \le H_3^{l-2} \text{Var}[T_i^{(s,3)}] l!/2$ for $H_3 \triangleq c_9 h^{\beta-d}$ and $l > 2$.
Thus, from Lemma \ref{lem_Bernstein_ind},
\vspace{-2mm}
\[
\mb P^{\otimes n} (\frac{1}{n}\sum_{i=1}^n{| T_i^{(s,1)}|} \geq \epsilon_1 ) \le 2 \exp (-\frac{n h^d \epsilon_1^2}{2(c_6+ c_5 \epsilon_1)})
\]
\vspace{-2mm}
\[
\mb P^{\otimes n} (\frac{1}{n}\sum_{i=1}^n|{ T_i^{(s,2)} - \mb{E} T_i^{(s,2)}|} \geq \epsilon_2 ) \le 2 \exp ( -\frac{n h^d \epsilon_2^2}{ 2(c_8 h^2 + c_7 \epsilon_2)})
\]
\vspace{-2mm}
\[
\mb P^{\otimes n} (\frac{1}{n}\sum_{i=1}^n|{ T_i^{(s,3)} - \mb{E} T_i^{(s,3)}|} \geq \epsilon_3 ) \le 2 \exp (-\frac{n h^d \epsilon_3^2}{2 (c_{10}  h^{2\beta} + c_9 h^{\beta}\epsilon_3 )}),
\]
for all $\epsilon_1, \epsilon_2, \epsilon_3 > 0$.
Moreover, $\mb{E}|T_i^{(s,3)}| \le c_{11} h^{\beta}$ for some $c_{11} > 0$, and \linebreak
$\frac{1}{n}\sum_{i=1}^n{\mb{E}|T_i^{(s,2)}|} \le O(h^{-d} \frac{1}{n}\sum_{i=1}^n{\|\eta^i_j - \eta\|_\infty}) \leq O(h^{-d} \frac{1}{n} \sum_{i=1}^n i^{-(\beta + d)/(2\beta + d)}) 
\leq O(h^{-d} \frac{1}{n} (1 + \int_{u=1}^n u^{-(\beta + d)/(2\beta + d)} du))  
\leq O(h^{-d} n^{-(\beta + d)/(2\beta + d)})  \leq c_{12} h^{\beta}$ 
for some $c_{12} > 0$ since $h = n^{- 1/(2\beta+d)}$.
Thus, we can obtain the new Eq. \eqref{eqn:bound3} as \linebreak
$\mb P^{\otimes n} \left ( |a_s| \geq \frac{c_1 \delta}{2M} \right ) \le 6 \exp(- c_{13} n h^d \delta^2)$ for some $C_6 > 0$ and $c_{13} > 0$.
And from the new Eq. \eqref{eqn:bound1}, \eqref{eqn:bound2}, and \eqref{eqn:bound3}, we can obtain Lemma \ref{lem:approX_jnd}.

\vspace{-1.6mm}

\bibliographystyle{splncs}
\bibliography{tamc15}

\end{document}